\documentclass{bmvc2k}

\usepackage{algorithm}
\usepackage{algpseudocode}
\usepackage{stfloats}

\title{Benchmarking Microsaccade Recognition with Event Cameras: A Novel Dataset and Evaluation}

\addauthor{Waseem Shariff*}{waseem.shariff@universityofgalway.ie}{1}
\addauthor{Timothy Hanley}{t.hanley4@universityofgalway.ie}{2}
\addauthor{Maciej Stec}{m.stec1@universityofgalway.ie}{1}
\addauthor{Hossein Javidnia}{hossein.javidnia@tcd.ie}{3}
\addauthor{Peter Corcoran}{peter.corcoran@universityofgalway.ie}{1}

\addinstitution{
C3I Imaging Lab, \\
School of Engineering, \\
University of Galway, \\
Ireland
}
\addinstitution{
School of Computer Science,\\
University of Galway, \\
Ireland
}
\addinstitution{
Department of Electronic and Electrical Engineering,
Trinity College Dublin, \\Ireland
}

\runninghead{Shariff et al.}{Microsaccade Recognition}

\begin{document}

\maketitle

\begin{abstract}
Microsaccades are small, involuntary eye movements essential for visual perception and neural processing. Traditional microsaccade research often relies on eye trackers and frame-based video analysis. While eye trackers offer high precision, they can be expensive and have limitations in scalability and temporal resolution compared to alternative methods. In contrast, event-based sensing offers a more efficient and precise alternative, as it captures high-resolution spatial and temporal information with minimal latency. This work introduces a pioneering event-based microsaccade dataset, designed to support efforts in studying small eye movement data within cognitive computing. Using Blender, we generate high-fidelity renderings of eye movement scenarios and simulate microsaccades with angular displacements ranging from 0.5° to 2.0°, divided into seven distinct classes. These sequences are simulated into event streams using v2e, preserving the temporal dynamics of real microsaccades. The event streams, with durations as small as 0.25 milliseconds and as large as 2.25 milliseconds, have such high temporal resolution, which prompted us to investigate whether SNNs could effectively detect and classify these movements. We evaluate the proposed dataset using Spiking-VGG11, Spiking-VGG13, and Spiking-VGG16, and further introduce Spiking-VGG16Flow - an optical flow - enhanced variant-implemented with SpikingJelly. Across experiments, these models achieve an average accuracy of approximately 90\%, effectively classifying microsaccades based on angular displacement, irrespective of event count or duration. These results highlight the suitability of SNNs for fine motion classification and establish a benchmark for future research in event-based vision. To facilitate further work in neuromorphic computing and visual neuroscience, both the dataset and trained models will be released publicly. The dataset and code will be made 
available here- \url{https://waseemshariff126.github.io/microsaccades/}
\end{abstract}

\maketitle

\section{Introduction}
\label{sec:introduction}


Microsaccades are small, involuntary eye movements typically under $1^\circ$ to $2^\circ$ in amplitude, distinct from larger, voluntary saccades due to their subtlety and small movements. While they share some kinematic properties with larger saccades, such as being conjugate in nature \cite{b3} and following a similar amplitude-velocity relationship \cite{b4}, microsaccades are much finer in movement and serve a different function. Predominantly binocular, they often exhibit a horizontal directional bias \cite{b7}. These brief, involuntary movements play a crucial role in maintaining visual clarity during fixation, subtly shifting the retinal image to prevent fading, a phenomenon that can occur when a stimulus remains stationary on the retina \cite{b1}. Unlike larger saccades, which are voluntary and are used for noticeable shifts in visual attention, microsaccades help preserve perceptual stability and visual sensitivity without any obvious change in gaze direction. The changes are very small, yet they are essential for preserving the clarity of our visual experience.

Microsaccades are attracting increased attention due to their sensitivity to underlying cognitive processes. Variations in their frequency, direction, and amplitude have been linked to attentional shifts, mental workload, and other internal neural states \cite{b1}. As brief, involuntary movements, they offer a noninvasive means of probing attention and cognitive function, making them relevant to both basic neuroscience and applied computational modeling \cite{b1,b2,b3}. They are also being explored as potential indicators for neurological and psychiatric conditions such as Parkinson’s disease, schizophrenia, and attention deficit disorders \cite{b7, b8, b9}. Despite this promise, microsaccades are difficult to analyze due to their small magnitude and transient nature. Current detection methods rely heavily on high-resolution eye trackers that produce numerical gaze coordinates, limiting the use of image- or event-based techniques. The absence of visual or spatiotemporal datasets tailored specifically for microsaccade detection further hinders the development of advanced models. In particular, it remains unclear whether artificial neural networks, especially those designed for vision tasks, can reliably detect and interpret such subtle motion signals amid noise and sparse data.

Despite the promise of microsaccades as indicators of cognitive and neurological states, their detection remains technically challenging \cite{b17}. Traditional frame-based cameras, such as standard RGB sensors, lack the temporal resolution needed to reliably capture these rapid, subtle eye movements, which occur on the order of milliseconds \cite{b18}. Achieving sufficiently high frame rates (e.g.,$\geq$1000 Hz) demands expensive hardware, large memory capacity, and generates vast amounts of mostly redundant data, complicating efficient analysis \cite{b18}. To address these limitations, event-based cameras also known as dynamic vision sensors (DVS) have emerged as a powerful alternative. Unlike conventional cameras that capture full image frames at fixed intervals, event cameras asynchronously record changes in brightness with microsecond-level precision and operate at effective sampling frequencies of around 10 kHz \cite{angelopoulos2020event, gallego2020event, shariff2024event}. This approach offers exceptionally high temporal resolution and data efficiency, making event cameras particularly well-suited for detecting rapid, subtle movements such as microsaccades. Previous studies have demonstrated the effectiveness of event cameras for high-speed eye tracking \cite{angelopoulos2020event, zhao2023ev, lenz2020event}, further motivating their use in this domain. 

Many event-based studies indicate that Spiking Neural Networks (SNNs) are particularly well-suited for processing event camera data due to their biological plausibility and dynamic neural behavior \cite{cordone2021learning, gallego2020event, shariff2024event}.However, while these models have demonstrated success in capturing spatiotemporal dynamics, it remains unclear whether they can reliably detect the small, brief eye movements characteristic of microsaccades, which typically last between 0.5 and 2.5 milliseconds. To investigate this, we frame the classification of eye motion angles based on angular bins, not to replicate biological function, but to evaluate whether neural networks can recognize such subtle motion patterns. This proof-of-concept task lays the groundwork for future, more complex objectives like trajectory reconstruction or cognitive state inference.

In this work, we introduce the first synthetic microsaccades dataset created using Blender. It comprises 175,000 annotated sequenced samples containing event streams for left-eye and right-eye microsaccades, over seven amplitudes ranging from 0.5° to 2°. This data set provides a controlled environment for systematic evaluation and supports reproducibility in neuromorphic vision.

We benchmarked the dataset using Spiking Neural Networks (SNNs), focusing on VGG-based models for their strong performance in visual tasks and compatibility with event-driven data \cite{nda}. By leveraging the publicly available repository \cite{nda}, we augmented the data and integrated event data with spiking models, building upon existing frameworks and improving the representation of the rapidly changing microsaccade signals. Additionally, we introduced Spiking-VGG16Flow, a variant of Spiking-VGG16 enhanced with optical flow estimation to improve sensitivity to subtle microsaccadic movements. This allowed a direct comparison against standard Spiking VGG models. Our classification results provide a solid baseline, demonstrating both the promise and the challenges of using SNNs for microsaccade detection.



\subsection*{Contributions}

\begin{itemize}
    \item We present the first synthetic event-based microsaccade dataset generated using Blender, featuring 175,000 annotated event streams that simulate realistic horizontal eye movements.
   
    \item We propose the Spiking-VGG16Flow variant, an optical-flow-augmented adaptation of Spiking-VGG16, specifically designed to investigate whether microsaccade classification models rely on event count or true motion cues.

    \item We demonstrate that models trained on synthetic data can generalize to real event-based recordings (EV-Eye), indicating the potential of simulation-driven learning in data-scarce domains.
    
\end{itemize}

\section{Background}
\subsection{Conventional Approaches to Microsaccade Detection and Analysis}

Microsaccades are critical for maintaining visual clarity during periods of visual fixation. They occur involuntarily and share kinematic similarities with larger, voluntary saccades, such as their conjugate nature and the amplitude-velocity relationship \cite{b2, b3, b4}. While microsaccades are often binocular, monocular events are typically excluded from analysis due to their perceived unreliability \cite{b14}. This approach is evident in studies such as Engbert and Mergenthaler (2006) and Troncoso et al. (2008), which recorded from both eyes but retained only binocular microsaccades in their analysis \cite{b15, b16}. This suggests that even if we record from just one eye, microsaccades can still be detected and classified, as long as they are part of a binocular event.

Despite their similarities, microsaccades differ significantly from larger saccades in terms of volitional control. Whereas saccades are often voluntary and goal-directed, microsaccades are considered involuntary. One of the strongest indicators of this involuntary nature is that microsaccadic onset times follow an exponential distribution, suggesting stochastic generation \cite{b6}. The operational definition of a microsaccade also varies slightly across studies, though most adopt a hard upper bound of $1^\circ$ 
 to $2^\circ$ for amplitude \cite{b7}. Historically, reported sizes ranged from as small as $10'$ \cite{b8} and $15'$ \cite{b9} to over $60'$ \cite{b10} in later studies. In this paper, we define microsaccades as having an amplitude of up to 2 degrees.

These findings were largely enabled by the use of high-frequency EyeLink eye tracking systems \cite{EyeLink}. Engbert and Kliegl (2003) relied on a system with a 250 Hz sampling rate \cite{b11}, while later studies, such as Laubrock et al. (2005) and Sinn and Engbert (2016), used the EyeLink II system at 500 Hz \cite{b13, b6}. More recent experiments by Krejtz et al. (2020) and Schneider et al. (2021) employed the EyeLink 1000 system, which offers a 1000 Hz sampling rate \cite{b2, b12}. These tools, along with velocity-based detection algorithms like the Engbert \& Kliegl method, have become the de facto standard for microsaccade analysis.

Efforts to move beyond hardware-based tracking are limited. For example, \cite{b17} used a CNN to segment the iris in 96 FPS video and detect microsaccades by thresholding eye velocity. However, microsaccades occur within a 2 ms window, requiring at least 500 FPS, so their approach lacks sufficient temporal precision. Computational methods for microsaccade detection are emerging but remain underexplored, and the lack of publicly available datasets hampers progress. This gap led to the creation of our own synthetic dataset using Blender for controlled microsaccadic simulation.

\subsection{Event-based Microsaccade Detection}

While numerous studies \cite{li2024gaze, ryan2021real, kang2023exploring, angelopoulos2020event, zhao2023ev, lenz2020event, zhao2023ev, ryan2023real} have explored eye detection and tracking using event-based cameras, microsaccade detection in this domain remains largely unexplored. To our knowledge, there has been no established work that systematically investigates microsaccades using event-cameras. Most recently, Iddrisu et al. (2024) underscored this research gap in their review, noting the lack of prior work on event-driven microsaccade detection within the context of event-based eye analysis \cite{b18}.



\section{Dataset and Preprocessing}
This section describes the creation of a synthetic microsaccade dataset using Blender-rendered RGB frames, followed by the conversion of these frames into event-based streams for microsaccade classification. The dataset is designed to provide controlled variations in motion and timing for model training and evaluation.

\subsection{Blender RGB Frame Generation}

To generate microsaccade data, we use Blender to render the RGB frames that simulate horizontal eye movements. Each microsaccade instance is modeled as a brief horizontal shift, where the eye moves in a random direction and then returns to its original position, mimicking the rapid and transient microsaccade movements typically observed during visual fixation. Movement is further defined in terms of angular displacement, and we established seven distinct classes for microsaccades, each corresponding to a specific angular range, as detailed in Table~\ref{tab:microsaccade_classes}. 

The left and right eyes exhibit symmetric motion patterns, so we first generate the sequences for the left eye. Afterward, these sequences are flipped to simulate the right eye. This approach captures the symmetrical nature of microsaccadic movements while maintaining consistency across both eyes. The motion of the eye is represented by a boomerang-like pattern, where the eye moves in one direction (e.g., leftward) and then returns to its original position (e.g., rightward), simulating the back-and-forth nature of a microsaccade.

\begin{figure*}[!h]
    \centering
    \includegraphics[width=1\textwidth]{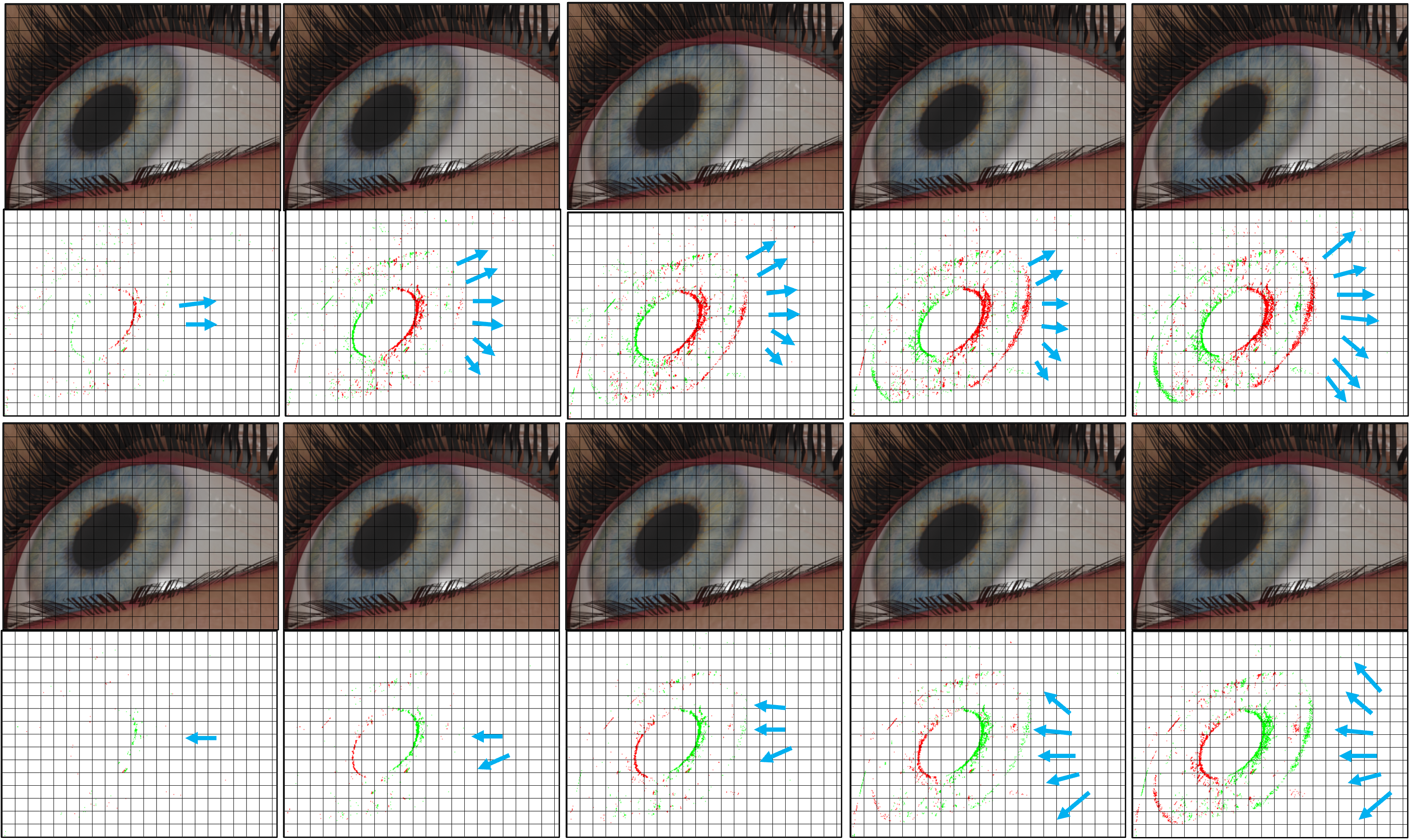}
    
    \caption{Visualization of a $1.0^\circ$ microsaccade sequence, showing Blender-rendered frames (top and bottom rows) and corresponding event streams (middle and fourth rows). A grid is overlaid for clarity, with red and green pixels representing positive and negative polarity events, respectively. All rows should be read from left to right to observe the temporal evolution of the microsaccade.}

    \label{fig:your_label}
\end{figure*}

In Blender, we simulate eye movements by rotating the virtual eyeball while keeping the camera fixed relative to the head. The eyeball’s rotation angles are carefully adjusted to replicate realistic microsaccadic velocities, producing motion consistent with natural fixational eye movements. The total number of frames generated for each class is proportional to the angular displacement, as shown in Table~\ref{tab:microsaccade_classes}. The class-specific frame counts are designed to accurately represent the motion duration for each class. To isolate amplitude as the sole distinguishing factor, we kept scene layout, lighting, textures, and background constant, enabling focused analysis of event responses to fine motion variations. A visualization of a $1.0^\circ$ microsaccade sequence is shown in Figure~\ref{fig:your_label}, where the first and third rows display the Blender-rendered frames cropped around the eye region, and the second and fourth rows show the corresponding event streams. Additional details of the Blender-based microsaccade data generation process are provided in the supplementary material.

\subsection{Event Simulation from RGB to Event}

To simulate event-based vision sensors, we convert the rendered RGB frames into event streams using an event simulator, specifically v2e \cite{v2e}. This conversion process mimics the asynchronous output of a Dynamic Vision Sensor (DVS), generating a series of events with timestamps, spatial coordinates $(x, y)$, and polarity values (increasing light intensity/decreasing light intensity). v2e has proven to be an effective tool for generating synthetic event-based data and is increasingly utilized in the evaluation of neuromorphic vision models. Numerous studies have employed v2e-generated datasets to demonstrate robust model performance and generalization across various tasks such as driver distraction \cite{yang2022event, shariff2023distraction}, Driver monitoring system \cite{ryan2023real}, MVSEC dataset \cite{v2e} and low-light image enhancement (LIE) dataset \cite{liang2024towards}

The time intervals for event generation are aligned with the time taken for each class, as presented in Table~\ref{tab:microsaccade_classes}. These intervals reflect the motion duration for each class, ranging from 0.25 ms for the smallest microsaccades (0.5°) to 2.25 ms for the largest microsaccades (2.0°). The duration is important, as it ensures that the event stream properly simulates the motion dynamics associated with different angular displacements, while overlapping these durations helps prevent the model from learning solely on the basis of timing cues, forcing it to focus on the underlying motion dynamics. To further enhance the simulation and mitigate bias from varying event counts across different classes, we also employ an event count overlapping strategy. Since event count is influenced by the magnitude of the motion, larger microsaccades typically produce more events. To address this, we randomly resample each event sequence after event simulation to introduce overlapping event counts across angular classes (more information in supplementary material). This ensures that each class cannot be solely distinguished by its event count while maintaining temporal consistency and producing an overall more balanced dataset. 

In total, each microsaccade class generates five distinct event sequences by randomly sampling durations within the specified time interval range for that class. These event sequences are resampled a further five times, as described in the supplementary material, expanding the dataset and ensuring that the model learns to classify based on motion dynamics rather than duration or event count. The resampling process further refines the dataset, resulting in a more robust set of event samples for model training.

Table~\ref{tab:dataset_details} summarizes the key properties of our synthetic microsaccade dataset\footnote{\url{https://waseemshariff126.github.io/microsaccades/}}. 
The original event frame resolution is $800\times600$ pixels, from which we extract a centered ROI of $440\times300$ pixels 
to focus on the eye region and reduce background noise. 
The dataset contains a total of 175{,}000 sequences, evenly split between the left and right eyes (87{,}500 each). 
We define seven horizontal microsaccade classes, with an equal number of samples per class to ensure balanced training and evaluation.

\begin{table}[!h]
\centering
\scriptsize
\begin{minipage}{0.45\linewidth}
\centering
\setlength{\tabcolsep}{4pt} 
\begin{tabular}{|c|c|c|c|}
\hline
\textbf{Class (°)} & \textbf{Frames} & \textbf{Angle (°)} & \textbf{Time (ms)} \\ \hline
0.5  & 7  & 0.000–0.625     & 0.25–0.75  \\ \hline
0.75 & 9  & 0.625–0.875     & 0.50–1.00  \\ \hline
1.0  & 11 & 0.875–1.125     & 0.75–1.25  \\ \hline
1.25 & 13 & 1.125–1.375     & 1.00–1.50  \\ \hline
1.5  & 15 & 1.375–1.625     & 1.25–1.75  \\ \hline
1.75 & 17 & 1.625–1.875     & 1.50–2.00  \\ \hline
2.0  & 19 & 1.875–2.125     & 1.75–2.25  \\ \hline
\end{tabular}
\caption{Microsaccade classes with corresponding frame counts, angle ranges, and temporal resolution. }
\label{tab:microsaccade_classes}
\end{minipage}
\hspace{0.05\linewidth} 
\begin{minipage}{0.45\linewidth}
\centering
\setlength{\tabcolsep}{4pt}
\begin{tabular}{|l|c|}
\hline
\textbf{Property} & \textbf{Value} \\
\hline
Original resolution & 800 $\times$ 600 \\
ROI resolution & 440 $\times$ 300 (center) \\
Left eye sequences & 87,500 \\
Right eye sequences & 87,500 \\
Total sequences & 175,000 \\
Number of classes & 7 \\
Class distribution & Evenly split \\
\hline
\end{tabular}
\caption{Synthetic microsaccade dataset details.}
\label{tab:dataset_details}
\end{minipage}
\end{table}



\section{Spiking VGG Architectures and Motion-Regularized VGGFlow}
\label{sec:vggflow_method}

We explore the problem of microsaccade classification using spiking convolutional architectures, beginning with spiking variants of VGG11, VGG13, and VGG16, adapted from~\cite{nda}. These models incorporate leaky integrate-and-fire (LIF) neurons and surrogate gradient training, providing strong baselines for classifying microsaccades of varying magnitudes in a neuromorphic setting.

During early experimentation, we observed a strong performance across classes, but it remained unclear whether the spiking models were truly learning the spatiotemporal dynamics of motion or simply leveraging differences in event count. Larger microsaccades naturally generate more events due to increased pixel activation, and we were concerned that this correlation might inadvertently drive classification accuracy. To investigate this hypothesis and ensure that the model's decision-making was not based on such statistical biases, we introduced an auxiliary motion-prediction objective into the spiking architecture.

Specifically, we propose \textit{spiking-VGGFlow}, a motion-regularized version of spiking VGG16 that incorporates an optical flow prediction branch during training. The model shares its convolutional backbone with spiking VGG16, but after the final convolutional block, it bifurcates into two heads: (1) a classification head composed of fully connected spiking layers trained with standard cross-entropy loss, and (2) a flow-prediction head that regresses dense optical flow fields between adjacent binned event frames.

This additional supervision encourages the network to learn motion-aware representations rather than overfitting to spatial event count. To compute flow targets, we first divide the event stream into temporally binned frames (e.g., 10 bins per 3 ms), then apply the Farneback algorithm~\cite{farneback2003two} on the polarity-summed frames to produce dense flow maps. Each target flow $F_t(x, y) = (u_t(x, y), v_t(x, y))$ represents horizontal and vertical displacements between frame $t$ and $t+1$. The flow prediction head minimizes the mean squared error between predicted and ground truth flow vectors:
\[
\mathcal{L}_{\text{flow}} = \frac{1}{HW} \sum_{x=1}^{W} \sum_{y=1}^{H} \left[ (\hat{u}(x, y) - u(x, y))^2 + (\hat{v}(x, y) - v(x, y))^2 \right].
\]
The overall training objective combines classification and flow supervision:
\[
\mathcal{L}_{\text{total}} = \mathcal{L}_{\text{class}} + \lambda \cdot \mathcal{L}_{\text{flow}},
\]
where $\lambda = 0.5$ balances the influence of motion consistency.

Importantly, the flow branch is only active during training. At inference time, the model reverts to the base spiking classifier without any added computational cost. As shown in Section~\ref{sec:results}, this motion-supervised training improves generalization and robustness, suggesting that the flow objective acts as an effective regularizer guiding the model toward more meaningful spatiotemporal representations and reducing reliance on event-count correlations.

\subsection{Training Setup}
The region of interest around the eye was cropped and processed at a resolution of 440×300 pixels. Models were implemented using the \texttt{SpikingJelly} \cite{fang2023spikingjelly} framework and trained for 30 epochs using the Adam optimizer with cosine annealing (step size: 10 epochs). The initial learning rate was 0.01, scaled with the batch size (default: 64), and weight decay set to $1 \times 10^{-4}$. We used a simulation length of 3 milliseconds discretized into 10 timesteps. All models were trained on our synthetic microsaccade dataset consisting of 7 angular classes, with optional Neuromorphic Data Augmentation (NDA)~\cite{nda}.

The dataset was split by eye, with separate models trained for left and right eye movements to account for spatial and appearance asymmetries between the two views. Although microsaccade dynamics are similar across eyes, differences in orientation, background, and illumination can introduce variability. However, per-eye training allows each model to specialize without having to learn invariance to these factors.  
Each subset contained 87,500 event streams, of which 20\% were set aside for validation. An independent test set consisting of 2,100 event streams was used to evaluate model generalization. Training was accelerated using dual NVIDIA A6000 GPUs. SpikingVGGFlow has approximately 15.9 million parameters, SpikingVGG16 about 14.7 million, SpikingVGG13 around 9.4 million, and SpikingVGG11 roughly 9.2 million parameters.






\section{Experiments and Results}
\label{sec:results}

\begin{table}[h!]
\centering
\resizebox{\textwidth}{!}{%
\begin{tabular}{l|c|c|c|c}
\hline
\textbf{Model} & \textbf{Accuracy (R/L)} & \textbf{Precision (R/L)} & \textbf{Recall (R/L)} & \textbf{F1-Score (R/L)} \\
\hline
\hline
\textbf{Spiking-VGG-Flow} & 91.81\% / 90.43\% & 0.922 / 0.912 & 0.918 / 0.904 & 0.918 / 0.904 \\
\textbf{Spiking-VGG16}\cite{nda}  & 92.10\% / 92.48\% & 0.923 / 0.933 & 0.921 / 0.925 & 0.920 / 0.924 \\
\textbf{Spiking-VGG13}\cite{nda}  & \textbf{95.10\%} / \textbf{93.05\%} & \textbf{0.951} / {0.936} & \textbf{0.951} / \textbf{0.930} & \textbf{0.951} / \textbf{0.930} \\
\textbf{Spiking-VGG11}\cite{nda} & 93.29\% / 90.48\% & 0.934 / \textbf{0.941} & 0.933 / 0.905 & 0.933 / 0.904 \\
\hline
\end{tabular}%
}
\caption{Model performance comparison for right and left eye results. Bold indicates the best value for each metric and eye.}
\label{tab:average_metrics}
\end{table}


Table~\ref{tab:average_metrics} summarizes the performance of 8 tested models on the microsaccade classification task. Among the architectures, VGG13 achieved the highest accuracy (93.05\%) and consistently strong precision, recall, and F1-scores, indicating a robust and well-balanced performance. VGG16 also performed competitively, while VGG11 and VGG16Flow achieved solid results with slightly lower overall metrics.

The VGG16Flow model, which incorporates optical flow supervision, was introduced to encourage the network to focus on motion dynamics rather than relying solely on event count, a factor that can vary with microsaccade magnitude. While VGG16Flow did not surpass VGG16 in overall accuracy, it delivered comparable recall (0.904 vs. VGG16’s 0.925) and demonstrated competitive performance across all metrics. These results suggest that flow supervision may offer subtle improvements in motion sensitivity, particularly in scenarios where directionality is more complex or difficult to distinguish.

Although flow regularization did not result in a dramatic performance boost, it proved to be a viable addition that maintained model effectiveness and may provide added value in tasks where fine-grained temporal distinctions are important. Overall, VGG16Flow presents a promising direction for enhancing spatiotemporal representations in event-based microsaccade classification.

\section{Evaluation on Real Events}
Since no public event-based datasets for microsaccade detection exist, we evaluated our model using the raw EV-Eye dataset \cite{zhao2023ev}, the only dataset offering long-duration, \textbf{raw} near-eye event streams. We focused on fixation periods, excluding normal saccades and blinks, as these are most likely to contain microsaccade-like activity. Despite the lack of ground truth labels for microsaccades, our model, trained on synthetic data, detected consistent patterns during fixation intervals. Using the VGGFlow model, which integrates spatial and event-flow information, we inference to show that the model can generalize to real-world data without retraining, highlighting proposed synthetic data's potential for training in data-scarce domains.

The table \ref{table:user_detection_performance} illustrates the model's ability to detect microsaccades across five users in the EV-Eye dataset. Despite the naturally low occurrence of microsaccades during sustained fixation, the model consistently identifies a small proportion of microsaccades, with detection rates ranging from 0.01\% to 0.63\%. This low detection rate reflects the inherent rarity of microsaccades in such conditions. The variation across users may stem from individual differences in oculomotor behavior, variations in recording quality, or noise within the event streams. Importantly, participants were not engaged in any cognitively demanding tasks. They were simply instructed to maintain fixation on a static point. Nevertheless, the model demonstrates strong generalization capabilities by detecting microsaccades without ground truth annotations and despite the domain shift from synthetic to real-world data. This underscores the potential of synthetic training data to support real-world event-based applications.

The successful detection of microsaccades under passive viewing conditions highlights the feasibility of using event-based sensors for subtle eye movement analysis. As future work, we plan to evaluate the model under more cognitively engaging scenarios to further explore its ability to capture task-related microsaccade patterns, which could provide deeper insights into attention and cognitive state estimation.

\begin{table}[ht]
\centering
\resizebox{\linewidth}{!}{%
\begin{tabular}{l|c|c|c}
\hline
\textbf{Subject} & \textbf{Total Number of Samples (3ms)} & \textbf{Microsaccades (\%)} & \textbf{Saccade/Fixation (\%)} \\
\hline\hline
user1 & 410,434  & 0.63  & 99.37 \\
user2 & 477,950  & 0.26  & 99.74 \\
user3 & 569,070  & 0.16  & 99.84 \\
user4 & 412,290  & 0.02  & 99.98 \\
user5 & 516,240  & 0.01  & 99.99 \\
\hline
\end{tabular}%
}
\caption{Detection performance of microsaccades and non-microsaccade events (saccades/fixations) across five users in the EV-Eye dataset, showing the total number of samples and the percentages classified as microsaccades (angular displacements between 0.5° and 2.0°) and non-microsaccade events.}
\label{table:user_detection_performance}
\end{table}

\section{Conclusion}

This work introduced the first synthetic event-based microsaccade dataset, developed using realistic simulations rendered in Blender. Given the suitability of event cameras for capturing rapid motion, these simulations were converted into event streams. The dataset was benchmarked using Spiking-VGG11, Spiking-VGG13, and Spiking-VGG16 to evaluate whether spiking neural networks (SNNs) can detect such brief and subtle eye movements. Although all models performed well, there was concern that predictions might rely more on event count than on motion characteristics. To address this, a variant called Spiking-VGG16Flow was proposed, incorporating an optical flow module to better capture motion across space and time. A dual-head architecture based on the Farneback framework was also explored, combining spiking and flow pathways to better interpret motion. Results showed that this model performed on par with standard SNNs, indicating that the original models were already learning from spatiotemporal motion patterns rather than simply relying on event count. All models accurately identified microsaccade classes based on angular displacement, regardless of event count or sample length. To assess generalization beyond synthetic data, Spiking-VGG16Flow was tested on the EV-Eye dataset. Despite the lack of labeled ground truth, the model consistently detected microsaccade-like activity during fixation periods, supporting the transferability of synthetic training to real-world conditions.

\textbf{Future Work.} A key limitation of this work is that the real event-camera data used was not collected under high cognitive load. To the best of our knowledge, there are currently no such datasets available. Despite this, our model was still able to detect small microsaccades with limited real data. A valuable next step would be to collect a dedicated event-based microsaccade dataset using a high-resolution eye tracker alongside an event camera, ideally including more cognitively demanding scenarios to better evaluate task-related microsaccade patterns. In addition, exploring the effect of different discretization strategies (e.g., varying bin counts, widths, or temporal resolutions) could further refine performance in temporal tasks such as microsaccade detection. Extending beyond our current seven-class formulation towards more complex objectives like trajectory reconstruction or start/offset detection may also help bring such models closer to real-world deployment.

\section*{Acknowledgment}
This work was supported in part by the Disruptive Technologies Innovation Fund (DTIF) under the EPICs project (Grant No. ‘DT 2023 0459A’). The authors would also like to thank Fotonation for their valuable participation in this study.

\bibliography{egbib}
\end{document}